%% file: arxiv.tex
\documentclass[10pt,letterpaper]{article}

\usepackage{arxiv/techreport}
\usepackage{afterpage}
\usepackage{placeins}
\newlength{\colw}\AtBeginDocument{\setlength{\colw}{0.62\textwidth}}
\newcommand{\figscale}[1]{1}
\newcommand{\releasestatement}{}
\newsavebox{\figboxbox}
\newcommand{\figbox}[2]{\sbox{\figboxbox}{#2}\typeout{FIGBOX natural width \the\wd\figboxbox}%
  \ifdim\wd\figboxbox>\linewidth\resizebox{\linewidth}{!}{\usebox{\figboxbox}}\else\usebox{\figboxbox}\fi}
\newlength{\colfigw}
\newcommand{\colfigbox}[2]{\sbox{\figboxbox}{#2}\setlength{\colfigw}{1.3\wd\figboxbox}%
  \ifdim\colfigw>\linewidth\setlength{\colfigw}{\linewidth}\fi\resizebox{\colfigw}{!}{\usebox{\figboxbox}}}
\newcommand{\firstmention}[1]{\textbf{#1}}
\input{preamble}
\definecolor{firstc}{HTML}{B9E4B0}
\definecolor{secondc}{HTML}{E3F4DE}
\renewcommand{\parahead}[1]{\par\addvspace{2pt plus 1pt}\noindent{\bfseries\sffamily #1}\hspace{0.6em}\ignorespaces}
\hypersetup{allcolors=reportblue,
  pdftitle={World Models as Native Latent Action Models},
  pdfauthor={Qinyue Li, Weijie Qiao, Qingyi Tao, Zerui Gong, Zhe Yang, Quanzhou Li, Taolin Zhang, Jinglin He, Zhonghua Wu, Chen Change Loy}}

\title{World Models as Native Latent Action Models}
\titleicon{\includegraphics[height=0.95em]{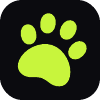}}
\teamlogo{\includegraphics[height=1.05cm]{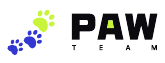}}
\author{%
  Qinyue~Li\aff{1*},
  Weijie~Qiao\aff{1,2*},
  Qingyi~Tao\aff{1*\textdagger},
  Zerui~Gong\aff{1},
  Zhe~Yang\aff{2},
  Quanzhou~Li\aff{2},
  Taolin~Zhang\aff{2},
  Jinglin~He\aff{1},
  Zhonghua~Wu\aff{1},
  Chen~Change~Loy\aff{2,3}}
\affiliations{\aff{1}SenseTime Research, Singapore,\enspace
  \aff{2}S-Lab, Nanyang Technological University,\enspace
  \aff{3}ACE Robotics}
\authornotes{\aff{*}Equal contribution,\enspace \aff{\textdagger}Project lead}
\keywords{Latent Action Models; Video World Models; Vision-Language-Action Models; Robot Learning.}
\reportlinks{%
  \reportlink{Project Page}{https://pawteam-eai.github.io/native-lam/}}

\begin{document}

\begin{titlecard}
\input{sections/00_abstract}
\end{titlecard}

\suppressfloats[t]
\begin{figure}[t]
\input{figures/teaser}
\teaserfigure
\end{figure}

\input{sections/01_introduction}
\input{sections/02_related_work}

\input{sections/04_method}
\input{sections/05_experiments}
\FloatBarrier
\input{sections/07_conclusion}

\input{references/bibliography}

\end{document}

%% file: preamble.tex
\usepackage[T1]{fontenc}
\usepackage{multirow}
\usepackage{makecell}
\usepackage{tikz}
\definecolor{firstc}{RGB}{233,157,157}
\definecolor{secondc}{RGB}{248,213,163}
\newcommand{\shadebox}[2]{{\setlength{\fboxsep}{1pt}\colorbox{#1}{#2}}}
\newcommand{\first}[1]{\shadebox{firstc}{\textbf{#1}}}
\newcommand{\second}[1]{\shadebox{secondc}{\underline{#1}}}
\newcommand{\firstbox}[1]{\first{#1}}
\newcommand{\secondbox}[1]{\second{#1}}
\definecolor{gradebase}{RGB}{242,150,60}
\colorlet{grade1}{gradebase!80}\colorlet{grade2}{gradebase!71}\colorlet{grade3}{gradebase!62}
\colorlet{grade4}{gradebase!54}\colorlet{grade5}{gradebase!46}\colorlet{grade6}{gradebase!38}
\colorlet{grade7}{gradebase!31}\colorlet{grade8}{gradebase!24}\colorlet{grade9}{gradebase!17}
\colorlet{grade10}{gradebase!10}

\usetikzlibrary{arrows.meta,positioning,patterns}
\graphicspath{{figures/}}

\definecolor{linkblue}{rgb}{0.21,0.49,0.74}
\usepackage[pagebackref,breaklinks,colorlinks,allcolors=linkblue]{hyperref}
\makeatletter
\AddToHook{begindocument/end}{%
  \let\nl@orig@citex\@citex
  \def\@citex[#1][#2]#3{%
    \edef\nl@keys{\zap@space#3 \@empty}%
    \@for\nl@key:=\nl@keys\do{%
      \ifx\nl@key\@empty\else\expandafter\Hy@backout\expandafter{\nl@key}\fi}%
    \nl@orig@citex[#1][#2]{#3}}}
\makeatother

\newcommand{\methodname}{Native LAM}
\newcommand{\policyname}{PAWlicy}
\newcommand{\policyfull}{\textbf{P}retraining \textbf{A}ctions from \textbf{W}orld models}

\newcommand{\latact}{\hat{\mathbf{a}}}
\newcommand{\vlat}{\mathbf{v}}
\newcommand{\reg}{\mathbf{r}}

\newcommand{\parahead}[1]{\par\addvspace{4pt plus 1pt minus 1pt}\noindent\textbf{#1}\hspace{0.8em}\ignorespaces}

%% file: sections/00_abstract.tex
\begin{abstract}
Large and diverse video sources, such as egocentric human video and recordings
from heterogeneous robots, carry rich dynamics but come either without actions
or without a shared action space. Latent action models (LAMs) can recover these
dynamics, but they train their inverse- and forward-dynamics models (IDM, FDM)
from scratch, trading off pixel reconstruction against latent capacity and
accuracy. Pretrained video world models natively generate dynamics and produce
high-fidelity subsequent frames, yet represent them densely, entangled across
the hidden states of every layer. Viewing the world model as a LAM, we show that
these dynamics can be extracted with minimal changes to the model.
\firstmention{\methodname}\ operates one pretrained video diffusion transformer as both IDM and
FDM, with learnable dynamics registers as the only causal bridge between the two, and
trains on counterfactual temporal pairs that differ in motion alone, so that the
registers distill a video's dynamics into a compact native latent action. We
further propose a pretraining paradigm for vision-language-action (VLA)
policies, \policyfull\ (\firstmention{\policyname}), in which \methodname\ serves as a
dynamics teacher on large-scale video; unlike world action models, it generates
video neither during training nor during inference. Native latent actions
outperform existing LAMs on the LARY benchmark in both action classification
and control regression, and replay the same motion on new robots, in new
scenes, and from human hands to robots, although no transfer objective is used
in training. With a 3.8B model,
\policyname\ reaches a 93.6\% average success rate on RoboTwin~2.0 trained with
domain-randomized demonstrations, the highest among VLAs. On RoboDojo, it
achieves competitive results on every dimension we evaluate, and its
Precision, evaluated by us, is on par with the leading entries of the public
leaderboard. On real robots, it improves over the $\pi_{0.5}$ baseline on both
tasks. Both stages are trained only on publicly available data on top
of a publicly released world model\releasestatement.
\end{abstract}

%% file: figures/teaser.tex
\newcommand{\regstrip}[4]{%
  \foreach \k in {0,...,7}
    \fill[#3, draw=#4, thick, rounded corners=1pt]
      (#1+0.19*\k, #2-0.17) rectangle ++(0.17,0.34);}
\newcommand{\teaserfigure}{%
\centering
\figbox{\figscale{0.88}\linewidth}{%
\includegraphics{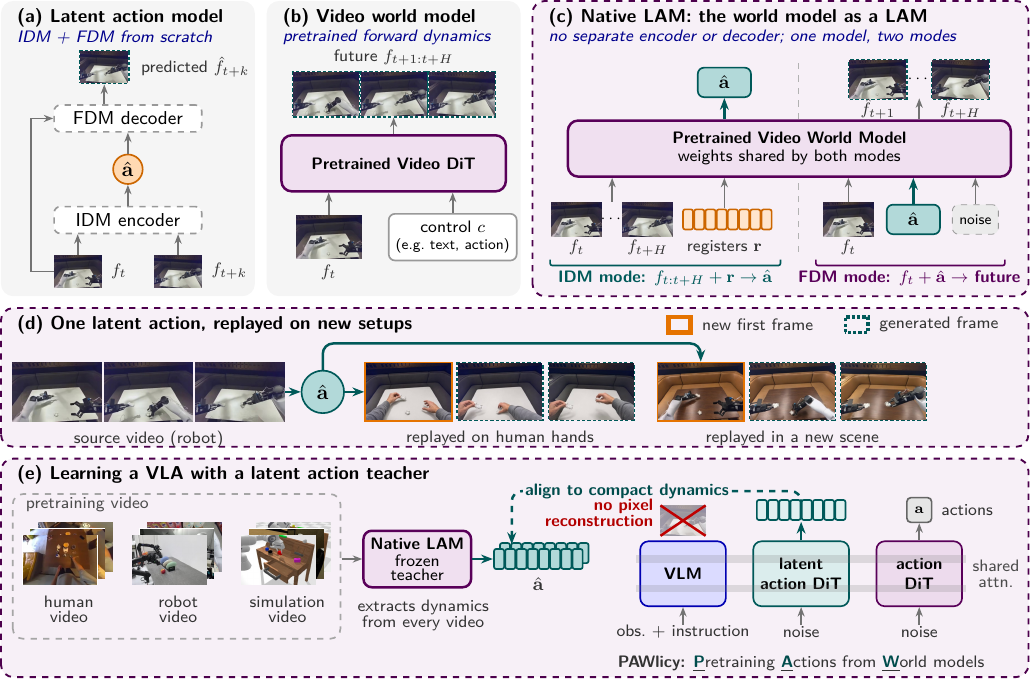}}
\captionof{figure}{\textbf{World models as native latent action models.}
(a) Existing LAMs learn an IDM and an FDM from scratch.
(b) A pretrained video world model is already an FDM.
(c) \methodname\ runs one world model in two modes, reading a latent action
$\latact$ from video and regenerating the future from $\latact$.
(d) Latent actions share one space and replay the same motion from edited first
frames in new setups.
(e) \policyname\ pretrains a VLA on $\latact$ without reconstructing pixels.}
\label{fig:teaser}
\vspace{0.2em}}

%% file: sections/01_introduction.tex
\section{Introduction}
\label{sec:introduction}

Video is the most abundant record of how agents act in the physical world,
spanning robots of many embodiments, human manipulation, and simulation. Much of
it cannot serve directly as action supervision, since egocentric human video
provides no robot action labels and robot datasets record actions in
embodiment-specific spaces. Latent action models
(LAMs)~\cite{schmidt2024lapo,bruce2024genie,ye2025lapa} address this gap. An
inverse-dynamics model (IDM) infers a latent action from an observed change, and
a forward-dynamics model (FDM) reproduces the change from it. The resulting shared
action space is used to pretrain vision-language-action (VLA)
policies~\cite{ye2025lapa,bu2025univla,chen2025villax,bjorck2025gr00t,li2026rotvla}
and to control world models~\cite{gao2025adaworld,gao2026dreamdojo,shen2026ld4wam}.

The quality of this space, however, is limited by how it is learned. A typical
LAM trains its IDM and FDM from scratch (\cref{fig:teaser}a), so the objective
trades off pixel reconstruction against the capacity and accuracy of the latent.
Training is also fragile~\cite{ye2025lapa,nikulin2025laom}. A loose bottleneck
can leak the future frame~\cite{schmidt2024lapo}, a tight one cannot describe the
motion~\cite{nikulin2025laom,ye2025lapa}, and in between the latent drifts toward
camera motion or distractors unrelated to the
agent~\cite{bu2025univla,nikulin2025laom}.

We start from a simple observation. A video world model is natively an FDM
(\cref{fig:teaser}b). Pretrained on large-scale robot and human video, it
carries a strong forward-dynamics prior on how scenes evolve, renders the future
faithfully, and implicitly captures inverse dynamics, which world action models (WAMs) exploit
to decode actions~\cite{li2025uva,zhu2025uwm,li2026lingbotva,yuan2026fastwam}. A
pretrained world model can thus be viewed as a unified LAM whose learning signal
need not be spent on rendering. Its dynamics, however, are entangled with
appearance across dense hidden states, with no compact representation from which
actions can be read. The challenge is to expose these dynamics natively through
a compact action interface.

\methodname\ (\cref{fig:teaser}c) appends a short, temporally ordered sequence
of learnable \emph{dynamics registers} to a pretrained video diffusion
transformer and runs it in two modes, with no separate encoder or decoder and
the registers as the only causal bridge between them. In \emph{IDM mode}, the
registers read a clean clip and yield the latent action. In \emph{FDM mode}, the
same network regenerates the future from the first frame and this latent alone,
with its original flow-matching objective and no text. Training on
\emph{counterfactual temporal pairs}, clips that share a first frame but differ
in motion, forces the latent to carry the motion. We term the resulting
representation \emph{native} latent actions, as it is produced and consumed by
the world model itself rather than an auxiliary encoder--decoder pair.
Native LAM thereby condenses the thousands of high-dimensional tokens that the
world model processes for a clip (3{,}456 tokens of dimension 2{,}560, using Kairos~3.1~\cite{kairos2026} as the
pretrained video DiT) into 32 latent
action tokens of dimension 128, without changing the network architecture or adding
a separate encoder or decoder.

We validate native latent actions in two respects. Quantitatively, they
outperform existing LAMs in both action classification and control regression
on the LARY benchmark (\cref{sec:exp-lary}). Qualitatively, a latent action extracted from
a robot video and replayed in FDM mode from a different first frame reproduces
the same motion in a new scene, on another robot, or with a human hand
(\cref{fig:teaser}d), although training uses no transfer objective.

Because they can be extracted from any video, native latent actions also enable
a new way to pretrain policies (\cref{fig:teaser}e). We propose \policyfull\
(\policyname), in which the frozen \methodname\ labels large-scale robot, human,
and simulated video, and a VLA learns to predict these latents and to act on
them. Unlike WAMs~\cite{li2025uva,zhu2025uwm,li2026lingbotva,yuan2026fastwam},
which make the policy generate appearance-dominated future video through a
multi-billion-parameter generator that is optimized during training and
sometimes also run at inference, \policyname\ keeps the world model's dynamics
and discards its rendering in both phases.
Teacher latents are computed once and cached, supervision is compact, and
inference runs a standard-size VLA. Both \methodname\ and \policyname\ are
trained only on publicly available datasets. On RoboTwin~2.0~\cite{chen2025robotwin2},
\policyname\ reaches a 93.6\% average success rate, the highest in
Tab.~\ref{tab:robotwin-sota}, and with clean demonstrations alone it still
outperforms $\pi_{0.5}$~\cite{black2025pi05} trained with the same past frame
and augmentation by 5.5 points. On RoboDojo~\cite{chen2026robodojo}, it leads
the compared methods on all four dimensions we evaluate, and its Precision,
evaluated by us, is on par with the leading entries of the public leaderboard. On a real robot, it lifts
$\pi_{0.5}$ from 23.3\% to 76.7\% on stacking three blocks and from 33.3\% to
90.0\% on putting shoes in a box.

In summary, our contributions are:
\begin{itemize}[leftmargin=*,itemsep=1pt,topsep=2pt]
  \item \methodname, in which a pretrained video world model serves as both IDM and FDM,
  bridged only by dynamics registers, without a separate encoder or decoder.
  \item A training recipe of counterfactual temporal pairs and a text-free FDM
  mode that makes the world model rely on its latent, applied to 46 robot,
  human, and simulation sources.
  \item \policyname, a VLA pretraining paradigm on native latent actions, with
  evidence that these latents are action-informative and transferable and
  improve over $\pi_{0.5}$ in simulation and the real world.
\end{itemize}

%% file: sections/02_related_work.tex
\section{Related Work}
\label{sec:related-work}

\parahead{Latent action models.}
LAPO~\cite{schmidt2024lapo} and Genie~\cite{bruce2024genie}
showed that an IDM/FDM pair trained to reconstruct future observations recovers
a latent action space from unlabeled video. Most later
LAMs~\cite{ye2025lapa,bu2025univla,chen2025villax,zhang2026dila,lee2026mvplam,zhang2026clap,gao2026dreamdojo,liu2026wla3}
follow this recipe and train their IDM and FDM from scratch. Nikulin
\etal~\cite{nikulin2025laom} show that, with distractors, the latents tend to
collapse onto pixel change.
Many LAMs further constrain the latent during
training. LAPA~\cite{ye2025lapa} and Moto~\cite{chen2025moto} quantize it into tokens,
RotVLA~\cite{li2026rotvla} models it as rotations, ALAM~\cite{tang2026alam}
regularizes it for composition and reversal consistency, and UniVLA~\cite{bu2025univla} conditions it
on language. villa-X~\cite{chen2025villax} adds proprioceptive supervision, which is closest to our
kinematic grounding.
Latent actions serve world models. AdaWorld~\cite{gao2025adaworld} and
DreamDojo~\cite{gao2026dreamdojo} use them as the control input of world models learned from
unlabeled video, and LD4WAM~\cite{shen2026ld4wam} learns latent dynamics from
human video and distills them from generated futures to condition a world
action model. In these methods the latent actions serve a world model but are
still extracted by a separate LAM. \methodname\ reverses this relation: the
world model itself extracts the latent action, and its own generator is the
decoder.

\parahead{World models for policy learning.}
Pretrained video generators such as Wan~\cite{wan2025wan} and robot-oriented
world models such as Kairos~\cite{kairos2026} encode strong dynamics priors, and
several lines of work bring them into policy learning. \emph{Video-as-policy}
methods generate a future and recover actions with an inverse-dynamics
model~\cite{du2023unipi,tian2025seer}. \emph{World action models} train a video
generator and an action generator jointly, so that one model predicts both
future frames and actions~\cite{li2025uva,zhu2025uwm,liao2026genie,cen2025worldvla,kim2026cosmospolicy,li2026lingbotva,shen2026ld4wam,bi2025motus,cai2026ahawam,chen2026lawam,luo2026beingh07};
Fast-WAM~\cite{yuan2026fastwam} shows that much of their benefit comes from video
co-training rather than test-time imagination; it keeps video generation as a
training objective and skips it at inference. \emph{Alignment} methods instead
supervise the policy, or its future tokens, with representations of future
observations---generated images~\cite{zhao2025cotvla}, video-model
features~\cite{hu2025vpp}, or frozen-encoder embeddings~\cite{zheng2025flare,zhang2025dreamvla,an2026awmvla,ma2026whatmatters,qiu2026agra}.
In all three, the policy is tied to dense visual targets that are dominated by
appearance, which recent studies identify as a source of brittleness under
visual shift~\cite{ma2026whatmatters,qiu2026agra}. \methodname\ uses the world
model differently: it distills the model's dynamics into a compact latent
action, and \policyname\ aligns to that latent alone, so the policy neither
renders pixels nor carries a video backbone.

\parahead{Visual representations for action.}
Self-supervised image and video encoders~\cite{simeoni2025dinov3,assran2025vjepa2}
are widely used as visual backbones for policies. LARY~\cite{nie2026lary}
probes such encoders and LAMs for semantic actions and low-level robot
control. Encoder features are dense, with hundreds of patch tokens per frame,
whereas a latent action is a compact bottleneck of a few tokens per segment.
We therefore compare with embodied LAMs under LARY's protocol and show that a
compact world-model-derived latent action is more action-informative than
existing ones.

\input{figures/method}

%% file: figures/method.tex
\begin{figure*}[t]
\centering
\figbox{\figscale{0.93}\linewidth}{%
\includegraphics{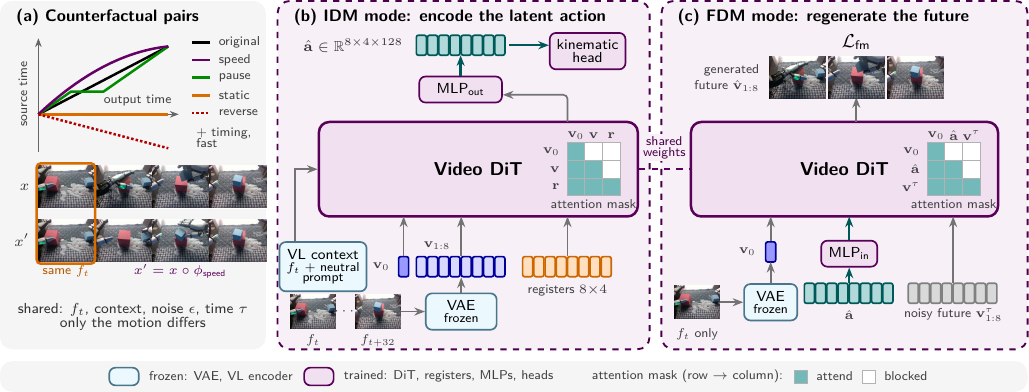}}
\caption{\textbf{Training \methodname.}
(a) Each video $x$ is paired with a retimed copy $x'=x\circ\phi$ of the same
footage that shares the first frame but not the motion.
(b) In IDM mode, dynamics registers read the clean video through a read-only
attention mask and are projected to the latent action $\latact$; on AgiBot World
and DROID, a kinematic head grounds $\latact$ in end-effector motion.
(c) In FDM mode, the same DiT regenerates the future from $\vlat_0$ and $\latact$
alone, with $\latact$ inserted into the token sequence through the adapter $\mathrm{MLP}_{\text{in}}$, taking over the conditioning role of text,
trained with its original flow-matching loss; both clips of a pair share the
noise and diffusion time, so only $\latact$ can explain their difference.}
\label{fig:method}
\end{figure*}

%% file: sections/04_method.tex
\section{Method}
\label{sec:method}

We first train \methodname, which turns a pretrained video world model into a
latent action model by running it in two modes linked by dynamics registers
(\cref{fig:method}; \cref{sec:native,sec:listen}). The trained \methodname\ then
serves as a frozen dynamics teacher for \policyname: it extracts native latent
actions from a mixture of robot, human, and simulated video, on which a VLA is pretrained and then
fine-tuned on target-robot demonstrations (\cref{sec:policy}).

\subsection{\methodname: one world model, two modes}
\label{sec:native}

\parahead{Preliminaries.}
From frames $f_t$ and $f_{t+k}$ without action labels, a LAM learns an
inverse-dynamics encoder $E$ and a forward-dynamics decoder $D$,
\begin{equation}
  \latact = E(f_t, f_{t+k}), \qquad
  \hat{f}_{t+k} = D(f_t, \latact),
  \label{eq:lam}
\end{equation}
trained so that $\hat{f}_{t+k}$ matches $f_{t+k}$, with a bottleneck on
$\latact$ that keeps it from copying the future.

\parahead{One backbone, two modes.}
\methodname\ has no separate encoder or decoder. Both $E$ and $D$ are the same
pretrained video diffusion transformer $G_\theta$, the publicly released
Kairos~3.1 4B robot world model~\cite{kairos2026}, which
operates on the latents of a frozen video VAE: a clip is encoded as
$\vlat=(\vlat_0,\vlat_1,\dots,\vlat_N)$, where $\vlat_0$ encodes the first frame
and each later $\vlat_i$ a short temporal segment. $G_\theta$ runs in an IDM
mode and an FDM mode with shared weights, and a short sequence of learnable
\emph{dynamics registers} $\reg$ is the only causal bridge between them
(\cref{fig:method}b--c).

\parahead{IDM mode: registers read the clip.}
We append $\reg$ to the tokens of a clean clip and run $G_\theta$ at $\tau=0$
(\cref{fig:method}b). Under a read-only attention mask (\cref{fig:method}b,
right), the registers attend to all video tokens and to one another, but no
video token attends to them, so the video stream is computed exactly as in the
pretrained model and the registers act purely as readers of its internal
states. A small output MLP (\cref{fig:method}b) maps their final
states to the latent action,
\begin{equation}
  \latact = E_\theta(\vlat_{0:N}) = \mathrm{MLP}_{\text{out}}\big(G_\theta(\vlat_{0:N}, \reg)_{\reg}\big),
  \label{eq:idm}
\end{equation}
where $(\cdot)_{\reg}$ denotes the outputs at the register positions;
$E_\theta$ takes the place of $E(f_t, f_{t+k})$ in \cref{eq:lam}. Grouped by time,
$\latact$ forms a short, temporally ordered sequence of continuous tokens.

\parahead{FDM mode: the world model replays the latent.}
The same network then generates the future from the first frame and the latent
action alone (\cref{fig:method}c), taking the place of $D(f_t, \latact)$:
\begin{equation}
  \hat{\vlat}_{1:N} = D_\theta(\vlat_0, \latact) = G_\theta\big(\vlat_0, \mathrm{MLP}_{\text{in}}(\latact)\big),
  \label{eq:fdm}
\end{equation}
where $G_\theta$ now generates by integrating its flow from noise, and the
adapter $\mathrm{MLP}_{\text{in}}$ maps $\latact$ to the width of $G_\theta$ and
replaces the text or action condition that the world model would otherwise
receive. The input is thus $[\vlat_0, \mathrm{MLP}_{\text{in}}(\latact), \vlat^{\tau}_{1:N}]$
(\cref{fig:method}c, bottom). FDM mode sees no text and no clean
future, so all information
about how the scene unfolds must be carried by $\latact$. We train it with the
backbone's own flow-matching objective on the future latents
(\cref{fig:method}c, top): with
$\boldsymbol{\epsilon}\sim\mathcal{N}(0,I)$, $\tau\in[0,1]$, and
$\vlat^{\tau}_{1:N}=(1-\tau)\vlat_{1:N}+\tau\boldsymbol{\epsilon}$,
\begin{equation}
  \mathcal{L}_{\text{fm}} =
  \mathbb{E}_{\tau,\boldsymbol{\epsilon}}\,
  w(\tau)\,\big\| u_\theta\big(\vlat^{\tau}_{1:N}, \tau \mid \vlat_0, \mathrm{MLP}_{\text{in}}(\latact)\big)
  - (\boldsymbol{\epsilon} - \vlat_{1:N}) \big\|^2 ,
  \label{eq:flow}
\end{equation}
where $u_\theta$ is the velocity predicted by $G_\theta$ and $w(\tau)$ the
backbone's timestep weighting~\cite{lipman2023flow,liu2023rectified}. Gradients
flow through the generator into $\latact$, the registers, and the shared
weights. Reading a latent action takes one forward pass on a clean clip, with no
denoising.

\subsection{Making the world model rely on the latent}
\label{sec:listen}

\parahead{Counterfactual temporal pairs.}
A pretrained world model that sees the first frame can already imagine a
plausible future. Trained directly on video from its own pretraining domain, it
suffers from posterior collapse: the FDM ignores the newly initialized latent
pathway and leaves the registers uninformative (\cref{sec:exp-collapse}). We remove this shortcut.
For each clip $x$ we draw a time map $\phi$ and build $x'=x\circ\phi$
from the same source video, with $x'_0=x_0$ (\cref{fig:method}a). The map $\phi$ holds
the first frame (\emph{static}), pauses mid-clip (\emph{pause}), changes speed
mid-clip (\emph{speed}), retimes the motion toward the same endpoint
(\emph{timing}), or plays backward into the preceding frames or faster when the source allows.
Frames are taken at native indices: nothing is
interpolated, padded, or extrapolated. Both clips are read independently in
IDM mode; in FDM mode they share the first frame, the context, the
noise $\boldsymbol{\epsilon}$, and $\tau$. The two generation problems are
identical except for the latent, so the only way to solve both is for $\latact$
and $\latact'$ to carry the difference in motion.

\parahead{Kinematic grounding.}
For AgiBot~World~2026~\cite{agibotworld2026} and DROID~\cite{khazatsky2024droid},
the only two sources whose action labels we use, a shallow kinematic head (\cref{fig:method}b, top right) reads $\latact$ and
predicts, for each group, end-effector
waypoints at several horizons after the segment start: base-frame translation,
rotation $\log(R_{\text{target}}R_{\text{start}}^\top)$, and gripper change.
Targets are derived from measured trajectories and recomputed after the time
map, so a static clip has zero motion. The loss $\mathcal{L}_{\text{kin}}$ has
weight $\lambda_{\text{kin}}=0.1$, warmed up over the first 2,000 updates so that the world
model shapes the registers first. All other sources are video only, without
action or text labels. The full objective is
\begin{equation}
  \mathcal{L}_{\text{LAM}} = \mathcal{L}_{\text{fm}}(x) + \mathcal{L}_{\text{fm}}(x')
  + \lambda_{\text{kin}}\,\mathbf{1}_{\text{kin}}\big[\mathcal{L}_{\text{kin}}(x)
  + \mathcal{L}_{\text{kin}}(x')\big],
\end{equation}
where $\mathbf{1}_{\text{kin}}$ indicates a clip from these two sources.

\subsection{\policyname: pretraining actions from world models}
\label{sec:policy}

\input{figures/pawlicy}

\policyname\ is a pretraining paradigm in which the trained \methodname\ acts as
a frozen dynamics teacher. The teacher assigns native latent actions to every
video, irrespective of whether actions were recorded, and a VLA is trained to
predict these latents and to condition its actions on them. We instantiate the
paradigm with three experts that share attention (\cref{fig:pawlicy}): a
vision-language model (VLM) that encodes the observation and instruction, a
\emph{latent action DiT} that predicts native latent actions, and an
\emph{action DiT} that generates motor commands.

\parahead{Reconstructing the past to predict the future.}
The VLM is pretrained for vision-language understanding, and its
representations are not trained to match the latent space of \methodname.
Predicting future latents directly from them asks the latent action DiT to map
into this space and to forecast dynamics that have not yet been observed, all
at once. We ease this by adding a past latent group to the target and a past
frame to the input. \methodname\ encodes eight groups per clip; we truncate the target
$\latact^*_t$ to the first four, which keeps the prediction horizon of
\policyname\ close to that of common VLAs: one group for the
segment that has just been observed and three for the segments that follow. In addition,
the VLM receives, besides the current image, a past frame from the beginning
of the observed segment. Recovering the
past group from the two images is an inverse-dynamics problem whose answer is
contained in the input. This reconstruction objective therefore provides a dense
and well-posed signal that encourages the VLM output to align with the latent
space, and lets the future groups, which are generated jointly with it,
continue from a latent anchored in what has already been observed. Beyond using the past frame for this reconstruction, the
latent action DiT must also predict reliably when no past frame is available,
as at the start of an episode. We therefore drop the past frame at random, and
additionally at the start of an episode during fine-tuning.

\parahead{Architecture.}
The latent action DiT generates $\latact_t$ from noise while attending to the
VLM tokens. The action DiT generates an action chunk $\mathbf{a}_{t:t+K}$ while
attending, at every layer, to the VLM tokens and to the keys and values of the
latent action DiT. Through this shared attention, the predicted latent
conditions every action layer and serves as a plan expressed in the
representation of the world model. Both experts are trained with flow
matching,
\begin{equation}
  \mathcal{L}_{\text{PAW}} =
  \lambda\,\mathcal{L}_{\text{fm}}\big(\latact_t;\,\latact^*_t\big)
  + \mathbf{1}_{\text{act}}\,\mathcal{L}_{\text{fm}}\big(\mathbf{a}_{t:t+K};\,\mathbf{a}^*_{t:t+K}\big),
  \label{eq:paw}
\end{equation}
where $\mathcal{L}_{\text{fm}}(y;\,y^*)$ is the flow-matching loss of \cref{eq:flow} for a prediction $y$ of the target $y^*$, and $\mathbf{1}_{\text{act}}$ indicates that action labels are used. Since every
video yields a latent target, egocentric human, cross-embodiment, and simulated
video all contribute to the dynamics pathway, whereas action-labeled robot data
additionally supervises the action DiT.\input{arxiv/sections/model_sizes}

\parahead{Pretraining and fine-tuning.}
We pretrain on this mixture and fine-tune on target-robot demonstrations with
the same objective, with the frozen teacher still providing latent targets.
\Cref{sec:exp-cost} compares the cost with world action models.

%% file: figures/pawlicy.tex
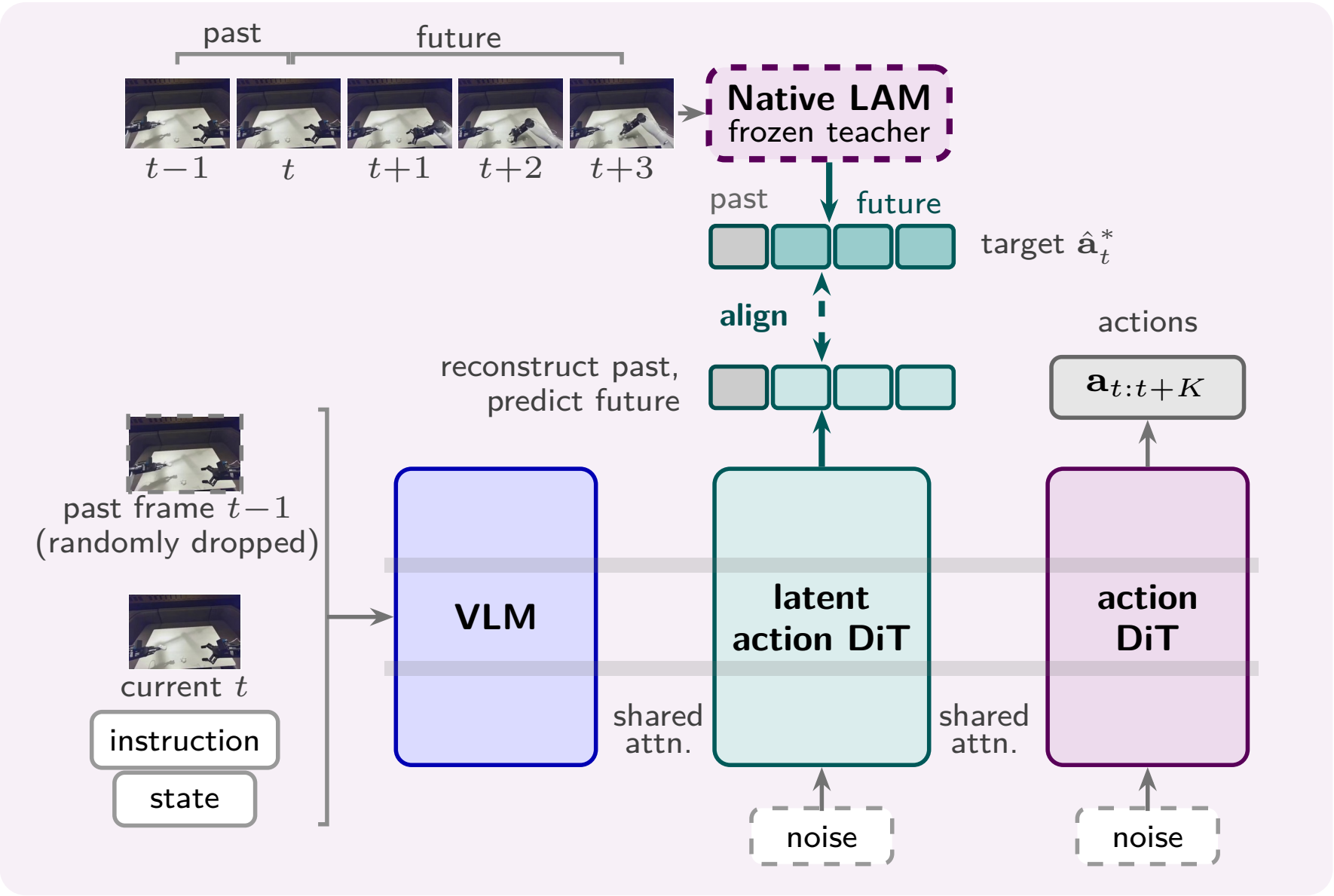
\begin{figure}[t]
  \centering
  \colfigbox{\figscale{0.97}\colw}{%
  \input{figures/pawlicy_body}}
  \caption{\textbf{\policyname.} A frozen \methodname\ labels each video window
  with one past and three future latent groups; $t{-}1, t, \dots, t{+}3$ index
  temporal groups, each containing four video frames. Sharing attention with the VLM,
  the latent action DiT reconstructs the past and predicts the future, and the
  action DiT conditions on it to generate actions. Video without actions still
  trains the latent pathway.}
  \label{fig:pawlicy}
\end{figure}

%% file: figures/pawlicy_body.tex
\includegraphics{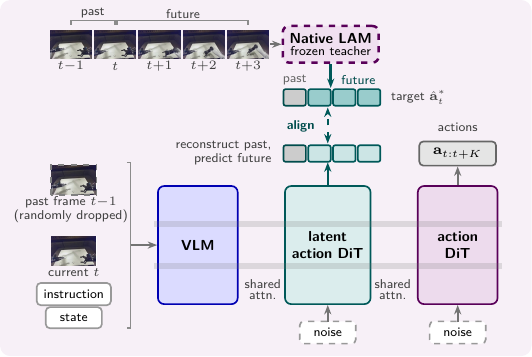}%

%% file: arxiv/sections/model_sizes.tex
\space The VLM is PaliGemma, initialized from the $\pi_{0.5}$ base
model~\cite{black2025pi05}. The latent action DiT and the action DiT
both have the architecture of the $\pi_{0.5}$ action expert, with 18 layers of
width 1024 (0.43B parameters each), so that \policyname\ has 3.78B parameters in total, against 3.35B for
$\pi_{0.5}$. At inference, both DiTs run 10 synchronous Euler flow-matching
steps; at each step, the latent action DiT updates $\latact_t$ and supplies its
per-layer keys and values to the action DiT.%

%% file: sections/05_experiments.tex
\section{Experiments}
\label{sec:experiments}

We evaluate what native latent actions capture and how they transfer
(\cref{sec:exp-analysis}), then test \policyname\ in simulation and on a real
robot and analyze its compute cost (\cref{sec:exp-policy}).

\subsection{Setup}
\label{sec:exp-setup}
\label{sec:data}
\input{arxiv/tables/sources}

\parahead{Implementation of \methodname.}
\methodname\ is trained on 46 video sources: real-robot datasets (65\% sampling
rate), human egocentric and hand-object video (30\%), and simulation (5\%). We
use main, head, or external cameras only. Starting from the public
backbone, we train for 13k iterations on 32 GPUs. Only AgiBot~World~2026 and
DROID provide kinematic grounding; all other data are video only, without action or text labels (\cref{tab:sources}). All human video comes from publicly released datasets, used under their licenses and the consent and de-identification procedures of their creators; we collect no new human data.

\parahead{Implementation of \policyname.}
\policyname\ is pretrained on 16 datasets: 12 of the \methodname\ sources and
four further datasets. The simulation
benchmarks RoboTwin~2.0 and RoboDojo are excluded from \policyname\
pretraining. The frozen
\methodname\ provides a latent target for every clip, and the action DiT is
supervised on the robot datasets marked in \cref{tab:sources}. For robot data, the
wrist-camera views of each episode are added to the VLM input. We pretrain \policyname\ for 50k iterations on 32 GPUs with a global batch size of 512.

\parahead{Benchmarks.}
\emph{LARY}~\cite{nie2026lary} probes representations for the robot actions
underlying a video. For closed-loop control, \emph{RoboTwin~2.0}~\cite{chen2025robotwin2}
provides 50 bimanual tasks with clean and domain-randomized scenes and tests
robustness to visual variation; \emph{RoboDojo}~\cite{chen2026robodojo}, a
recent benchmark with a public leaderboard, groups 42 simulation tasks into five
capability dimensions; and two tasks on a real AgileX NERO robot, stacking
three blocks and putting shoes in a box, test transfer to a physical
embodiment.

\parahead{Baselines.}
We compare against $\pi_{0.5}$~\cite{black2025pi05} and against $\pi_{0.5}$
with the same past frame and augmentation as \policyname\ (``$\pi_{0.5}$ +
past + aug.''; we use the online image augmentation of
PatchWAM~\cite{wang2026patchwam}), post-trained with the same end-effector action space, data,
and budget, so that the two differ only in the \policyname\ pretraining stage.

\subsection{\methodname\ as a latent action model}
\label{sec:exp-analysis}

\subsubsection{Action evaluation on LARY}
\label{sec:exp-lary}
\input{arxiv/tables/lary_float}

LARY~\cite{nie2026lary} probes a frozen representation of a video for the
actions it depicts, on two tracks: semantic action classification on composite
human and robot sets, and regression of the underlying end-effector actions on
CALVIN~\cite{mees2022calvin}, VLABench~\cite{zhang2024vlabench},
RoboCOIN~\cite{wu2025robocoin}, and AgiBot-Beta~\cite{bu2025agibot}. For the
regression track we encode the source video segment of each sample, since
\methodname\ reads clips rather than frame pairs.

\Cref{tab:lary} shows that \methodname\ is the strongest latent action model on
both tracks. It reaches 71.77\% classification accuracy and 0.25 regression
error on average, against 70.75\% and 0.37 for WLA$^3$, the best previous LAM. The margin holds on every dataset, and is largest
on the real-robot sets RoboCOIN (0.36 against 0.55) and AgiBot-Beta (0.41
against 0.57). None of the four regression datasets is used to train \methodname: AgiBot~World~2026
is collected on the G2 rather than the G1 of AgiBot-Beta, so the two share no
trajectories, and the other three are unseen. \methodname\ stays frozen during probing.

\subsubsection{Motion that transfers and interpolates}
\label{sec:exp-transfer}
\input{figures/transfer}

During training, each latent is decoded only from the first frame of its own
video. At test time, we read a latent from one video in IDM mode and replay it
in FDM mode from a different first frame (\cref{fig:transfer}), changing the
embodiment (a), the scene (b), the agent from a human to a robot hand (c), or
the domain from simulation to a photorealistic scene (d). In every case the
generated video performs the source motion with the new agent in the new
scene: the latent captures \emph{what} is done, largely independently of
\emph{who} performs it and \emph{in which scene}. 

\input{figures/interp}

Interpolation tests whether the latent space is organized by motion or by
appearance. We replay
$(1-\beta)\latact_{\text{still}}+\beta\latact_{\text{full}}$, mixing the
latents of a still clip and a full-action clip, from the same first frame
(\cref{fig:interp}). Each intermediate $\beta$ yields one coherent motion that
travels part of the way, further as $\beta$ grows, whereas blending the two
videos' pixels superimposes static and moving arms. A mixture of latents is
thus an intermediate \emph{motion}, not an intermediate image.
\input{arxiv/sections/gen_compare}

\input{arxiv/sections/collapse}

\subsection{\policyname\ in simulation and the real world}
\label{sec:exp-policy}
\input{arxiv/tables/robotwin_realrobot}\input{arxiv/tables/robotwin_sota_float}

\parahead{Simulation.}
On RoboTwin~2.0, we train on clean demonstrations only (50 per task) or on clean
and randomized demonstrations (50+500 per task), test on clean (c) or
randomized (r) scenes, giving c2c, c2r, r2c, and r2r, and run the official 100
evaluation episodes per task. \Cref{tab:robotwin} measures the effect of \policyname\ pretraining.
Trained on clean demonstrations only, \policyname\ improves over $\pi_{0.5}$
with the same past frame and augmentation by 4.18 points in clean scenes and by
6.78 points in randomized scenes, reaching 74.02\% on average against 68.54\%.
The gain is largest when the test scenes differ from the training scenes,
consistent with a latent that encodes motion rather than appearance.

With randomized demonstrations added, \cref{tab:robotwin-sota} places
\policyname\ among published results. It reaches 93.8\% in clean and 93.4\% in
randomized scenes; the clean-scene result is second only to
LD4WAM~\cite{shen2026ld4wam} (94.0\%), and both the randomized-scene result and
the average of 93.6\% are the best in the table. The average exceeds the
strongest VLA and world action model in the table and improves on RotVLA~\cite{li2026rotvla}, the
best previous VLA that uses latent actions, by 4.5 points, although
\policyname\ generates no video in training or inference.

On RoboDojo, we do not compare on Memory, which requires a memory module that
\policyname\ does not have.\input{arxiv/sections/memory} Following the official setting, we fine-tune on
its 3,500 released trajectories, 100 for each of the 34 training tasks plus 100
auxiliary ones (the Open tasks have no training data), and report success
rates over 50 episodes per task, averaged over three seeds; as \policyname\ is not on the leaderboard, we
evaluate it ourselves with the released code. \policyname\ is the best among the methods in
\cref{tab:robotwin-sota} on all four dimensions despite its relatively small
model size. The leading leaderboard entries~\cite{xsquare2026walloss,simate2026beta,rexm12026} build on in-house or undisclosed
pretraining data, whereas \policyname\ uses only public data; still, its
Precision is on par with the leading entries.

\parahead{Real world.}
We evaluate the two tasks in \cref{tab:realrobot}, fine-tuning each policy on
100 demonstrations per task and evaluating it over 30 trials. The robot is an AgileX NERO with two 7-DoF arms, observed through a head camera and two wrist cameras at 20\,Hz. The three policies are fine-tuned with the same recipe, separately on each task, and deployed alike: each call returns a 50-step action chunk that is executed in full before the next call. \policyname\ reaches 76.7\% and 90.0\%, against 66.7\% and 80.0\% for
$\pi_{0.5}$ with the same past frame and augmentation and 23.3\% and 33.3\% for
$\pi_{0.5}$. Similar to its advantage in Precision in simulation, \policyname\
performs more precise placement in both tasks. In stacking, \policyname\ places the cubes more precisely. In the shoe task, the failures of all three policies look alike: in most of them the right shoe does not end up flat in the box, but leans on the wall, tips onto the left shoe, or stands on its toe. \policyname\ fails in 3 of 30 trials, against 6 for $\pi_{0.5}$ with the same past frame and augmentation and 20 for $\pi_{0.5}$.

\input{arxiv/tables/efficiency}
\parahead{Training and inference cost.}
\label{sec:exp-cost}
World action models learn dynamics through a multi-billion-parameter video
expert, such as Wan~2.2 5B in Motus~\cite{bi2025motus} and
Fast-WAM~\cite{yuan2026fastwam} or Kairos~3.1 in Kairos-WAM~\cite{kairos2026}, whereas \policyname\ predicts native latent
actions with a 0.43B latent action DiT trained from scratch. On the 33-frame,
$384{\times}256$ clips of \methodname, processing 32 latent tokens instead of
the 864 video tokens of Wan~2.2 or the 3{,}456 of Kairos~3.1 reduces the training compute of the dynamics module by about $387\times$ and
$1{,}270\times$ (0.0605 against ${\approx}23.4$ and ${\approx}76.9$ TFLOPs per step at
batch size 1); both figures count each module
alone, and joint attention with the other experts adds cost on both sides. At inference on an RTX 6000 Ada, our deployment GPU, with all methods timed by
us, \policyname\ takes 106\,ms per step, on par with Fast-WAM without video
generation (103\,ms, no VLM), $1.8\times$ faster than Kairos-WAM, and
$5.2\times$, $5.1\times$, and $8.5\times$ faster than Fast-WAM (IDM, joint)
and Motus, which generate video at test time. \Cref{tab:efficiency} lists all measurements.

%% file: arxiv/tables/sources.tex
\begin{table}[t]
  \centering
  \caption{\textbf{Data sources and their supervision} in the two stages.
  \methodname\ is trained on video, and on end-effector actions for kinematic
  grounding where marked; \policyname\ is pretrained on the teacher's latent
  actions and, where marked, on robot actions. The sources below the line are
  used for \policyname\ only: \methodname\ never sees them, and they are
  labeled by the frozen \methodname. RoboTwin~2.0 and RoboDojo are excluded
  from \policyname\ pretraining.}
  \label{tab:sources}
  \small
  \setlength{\tabcolsep}{4pt}
  \begin{tabular}{@{}l >{\raggedright\arraybackslash}p{0.5\linewidth} cc cc@{}}
    \toprule
    & & \multicolumn{2}{c}{\methodname} & \multicolumn{2}{c}{\policyname} \\
    \cmidrule(lr){3-4}\cmidrule(l){5-6}
    Group & Sources & Video & Action & Latent action & Action \\
    \midrule
    Real robots & AgiBot~World~2026~\cite{agibotworld2026}, DROID~\cite{khazatsky2024droid} & \checkmark & \checkmark & \checkmark & \checkmark \\
    Real robots & ALOHA (static)~\cite{zhao2023aloha}, Bridge~V2~\cite{walke2023bridgev2},
      Galaxea~\cite{jiang2025galaxea}, Unitree G1~\cite{unitree2025datasets}, Unitree Z1~\cite{unitree2025datasets} & \checkmark & & \checkmark & \checkmark \\
    Real robots & Humanoid Everyday~\cite{zhao2025humanoideveryday}, MIME~\cite{sharma2018mime},
      1X World-Model Raw Data~\cite{onex2024worldmodel},
      Fourier ActionNet~\cite{fourier2025actionnet} & \checkmark & & \checkmark & \\
    Real robots & Fanuc Manipulation~\cite{zhu2023fanuc},
      GR00T Teleop G1~\cite{nvidia2025grootteleop}, GR00T Teleop GR1~\cite{nvidia2025grootteleop}, Jaco Play~\cite{dass2023jacoplay},
      1X 1xgpt~\cite{onex2024worldmodel},
      RoboChallenge~\cite{yakefu2025robochallenge}, RoboVQA~\cite{sermanet2024robovqa},
      TACO Play~\cite{rosete2022tacoplay},
      and OXE~\cite{oxe2024} subsets (Berkeley Autolab UR5~\cite{chen2023berkeleyur5},
      Stanford Kuka~\cite{lee2019stanfordkuka}, CMU Playing with Food~\cite{sawhney2021playingfood},
      CMU Play Fusion~\cite{chen2023playfusion}, CMU Stretch~\cite{bahl2023affordances,mendonca2023stretch},
      RoboSet~\cite{bharadhwaj2024roboagent}, Language Table~\cite{lynch2023languagetable}) & \checkmark & & & \\
    Human & Xperience-10M~\cite{ropedia2026xperience} & \checkmark & & \checkmark & \\
    Human & CASTLE~2024~\cite{rossetto2025castle},
      DexYCB~\cite{chao2021dexycb}, EgoDex~\cite{hoque2025egodex}, Ego-Exo4D~\cite{grauman2024egoexo4d},
      EgoExoLearn~\cite{huang2024egoexolearn}, EgoHands~\cite{bambach2015egohands}, EgoMe~\cite{qiu2025egome},
      EgoVid-5M~\cite{wang2025egovid}, EPIC-KITCHENS-100~\cite{damen2022epic100},
      FPVSum~\cite{ho2018fpvsum}, GigaHands~\cite{fu2025gigahands}, HOI4D~\cite{liu2022hoi4d},
      HoloAssist~\cite{wang2023holoassist}, Kinetics-400~\cite{kay2017kinetics},
      Motion-X++~\cite{zhang2025motionxpp}, OakInk2~\cite{zhan2024oakink2}, TACO~\cite{liu2024taco} & \checkmark & & & \\
    Simulation & RoboCasa~\cite{nasiriany2024robocasa}, RoboTwin~2.0~\cite{chen2025robotwin2} & \checkmark & & & \\
    \midrule
    Real robots & RoboCOIN~\cite{wu2025robocoin}, RoboMIND~\cite{wu2025robomind} & & & \checkmark & \checkmark \\
    Real robots & InternData-A1~\cite{tian2026interndata} (physical/A2D subset) & & & \checkmark & \\
    Simulation & CALVIN~\cite{mees2022calvin} & & & \checkmark & \checkmark \\
    \bottomrule
  \end{tabular}
\end{table}

%% file: arxiv/tables/lary_float.tex
\begin{table}[t]
  \centering
  \caption{\textbf{Evaluation on LARY}~\cite{nie2026lary} against embodied
  latent action models: action classification (top-1 \%) and control regression
  (normalized MSE). Baselines are taken from LARY and WLA$^3$~\cite{liu2026wla3}.
  $^\ast$See \cref{sec:exp-lary}. Best in
  \firstbox{bold}, second \secondbox{underlined}.}
  \label{tab:lary}
  \small\setlength{\tabcolsep}{3pt}
  \begin{tabular*}{\linewidth}{@{\extracolsep{\fill}}l ccc ccccc@{}}
    \toprule
    & \multicolumn{3}{c}{Classification (top-1 \%) $\uparrow$}
    & \multicolumn{5}{c}{Control regression (MSE) $\downarrow$} \\
    \cmidrule(lr){2-4}\cmidrule(l){5-9}
    Method & Comp.\ Human & Comp.\ Robot & Avg. & CALVIN & VLABench & RoboCOIN & AgiBot-Beta$^\ast$ & Avg. \\
    \midrule
    LAPA~\cite{ye2025lapa}            & 14.61 & 23.64 & 19.13 & 0.96 & 0.95 & 0.96 & 1.00 & 0.97 \\
    UniVLA~\cite{bu2025univla}        & 19.08 & 18.56 & 18.82 & 0.82 & 0.74 & 0.94 & 0.97 & 0.87 \\
    villa-X~\cite{chen2025villax}     & 17.80 & 29.90 & 23.85 & 0.86 & 0.72 & 0.94 & 0.97 & 0.87 \\
    DiLA~\cite{zhang2026dila}            & 64.41 & 62.10 & 63.26 & 0.55 & 0.27 & 0.86 & 0.82 & 0.63 \\
    MVP-LAM~\cite{lee2026mvplam}         & 50.31 & 51.61 & 50.96 & 0.55 & 0.38 & 0.90 & 0.92 & 0.69 \\
    CLAP~\cite{zhang2026clap}            & 55.32 & 63.02 & 59.17 & 0.43 & 0.17 & 0.68 & 0.60 & 0.47 \\
    DreamDojo~\cite{gao2026dreamdojo}    & 63.69 & 59.41 & 61.55 & 0.35 & 0.12 & 0.74 & 0.81 & 0.51 \\
    WLA$^3$~\cite{liu2026wla3}                  & \second{73.12} & \second{68.37} & \second{70.75} & \second{0.25} & \second{0.09} & \second{0.55} & \second{0.57} & \second{0.37} \\
    \midrule
    \methodname\ (ours)                         & \first{74.76} & \first{68.77} & \first{71.77} & \first{0.18} & \first{0.06} & \first{0.36} & \first{0.41} & \first{0.25} \\
    \bottomrule
  \end{tabular*}
\end{table}

%% file: figures/transfer.tex
\begin{figure*}[t]
  \centering
  \figbox{\linewidth}{%
  \includegraphics{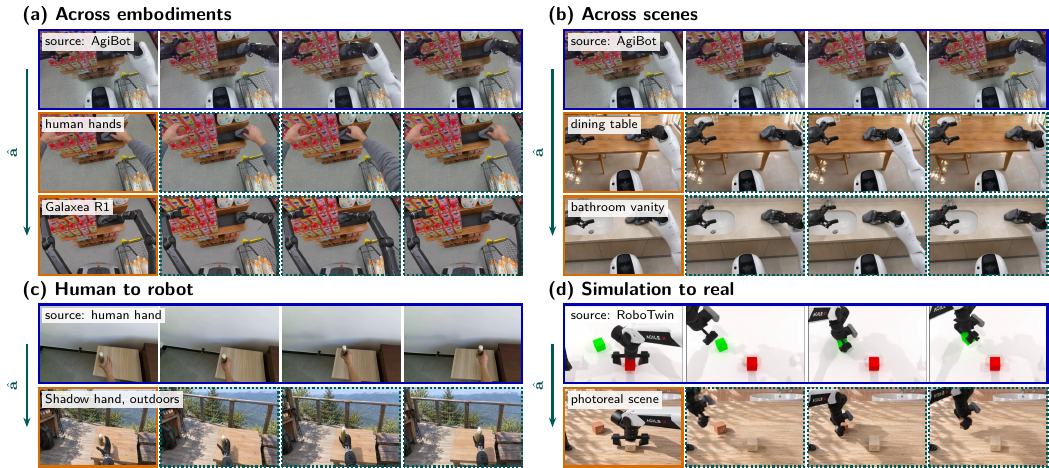}}
  \caption{\textbf{Native latent actions transfer.} A latent $\latact$ is read
  once from a source clip (blue) in the IDM mode of \methodname\ and replayed
  in its FDM mode from a different first frame (orange), which is generated
  with GPT image generation from the source first frame; dotted frames are
  generated. The generated videos carry out the source
  motion with another embodiment, in another scene, from human to robot, and
  from simulation to a photorealistic scene, although no transfer objective is used in
  training.}
  \label{fig:transfer}
\end{figure*}

%% file: figures/interp.tex
\begin{figure}[t]
  \centering
  \colfigbox{\colw}{%
  \includegraphics{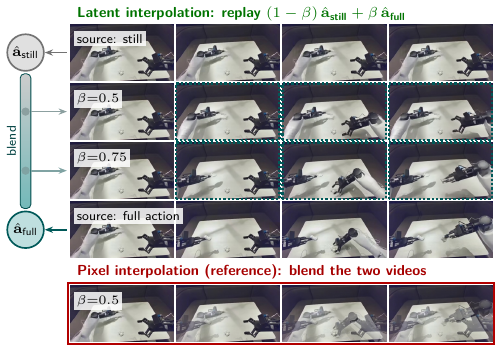}}
  \caption{\textbf{Interpolating latents interpolates the motion.} Latents read
  from a still clip and a full-action clip are blended and replayed from the same
  first frame. The top and bottom rows are the two source clips; the dotted
  frames between them are generated. Each intermediate latent yields one
  coherent motion that travels part of the way, further as $\beta$ grows;
  blending pixels instead superimposes static and moving arms (red box).}
  \label{fig:interp}
\end{figure}

%% file: arxiv/sections/gen_compare.tex
\subsubsection{Visual comparison with other LAMs}
\label{sec:exp-gen}
\input{figures/gen_compare_caption}
\begin{figure}[t]
  \centering
  \def\gccw{1.8}\input{figures/gen_compare_body}
  \caption{\gencomparecaption}
  \label{fig:gen-compare}
\end{figure}
We compare \methodname\ with four other LAMs when the latent actions
are decoded back into the clip they were read from, frame by frame and for the
whole clip. Frame by frame, the latents of the other LAMs reconstruct large motions poorly
in pixels: the arm and the banana blur or vanish during the fast transport
(third frame of \cref{fig:gen-compare}a). \methodname\
reconstructs both large and small motions sharply, which we attribute to the
prior of its pretrained world model, and the generated motion follows the
source. For the whole clip (\cref{fig:gen-compare}b), the frame-pair LAMs must roll out autoregressively,
so their errors accumulate and within a few steps the arm and the banana blur
away or stop moving. \methodname\ generates the clip in one pass, which these
LAMs cannot do directly, and keeps the arm, the banana, and the motion intact
to the end.

%% file: figures/gen_compare_caption.tex
\providecommand{\gencomparecaption}{\textbf{Frames generated from the latent
  actions of different LAMs.} A real-robot clip of picking up a banana and
  placing it in a basket (top row) runs left to right;
  the first frame is real.
  (a)~Frame by frame: every method, \methodname\ included, is used as a
  conventional LAM on frame pairs; for each two consecutive frames, it encodes
  the latent action between them (IDM) and generates the second frame from the
  real first one with its own world model (FDM), similar to teacher forcing.
  (b)~Whole clip (autoregressive): the other LAMs generate each frame from the
  previously \emph{generated} one with the latent action of the corresponding
  pair of real frames;
  \methodname\ reads the whole clip at once in IDM mode and generates the
  whole clip from the real first frame in one pass in FDM mode.}

%% file: figures/gen_compare_body.tex
\includegraphics{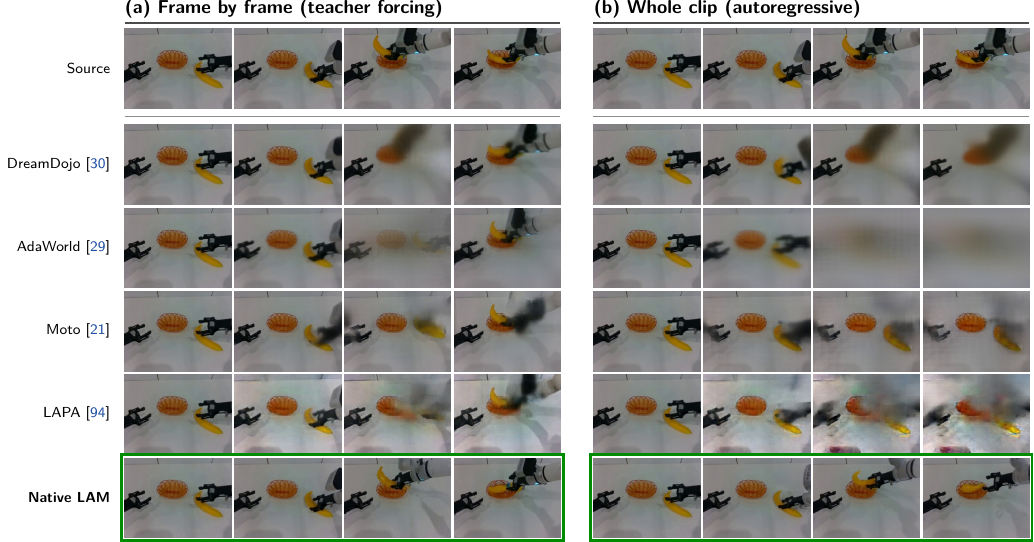}\nocite{gao2026dreamdojo,gao2025adaworld,chen2025moto,ye2025lapa}\csname Hy@backout\endcsname{gao2026dreamdojo}\csname Hy@backout\endcsname{gao2025adaworld}\csname Hy@backout\endcsname{chen2025moto}\csname Hy@backout\endcsname{ye2025lapa}

%% file: arxiv/sections/collapse.tex
\input{figures/supp_collapse}
\subsubsection{Counterfactual pairs prevent posterior collapse}
\label{sec:exp-collapse}
Without counterfactual pairs, the pretrained FDM learns to ignore the latent
(\cref{sec:listen}). $G_\theta$ is initialized from Kairos~3.1, pretrained on
large-scale robot and human manipulation video, and already predicts canonical
trajectories from $\vlat_0$ alone, while the dynamics registers $\reg$, the
output MLP, and the adapter $\mathrm{MLP}_{\text{in}}(\latact)$ start from
random weights. Optimization therefore suppresses the adapter pathway instead of
routing dynamics through $\reg$; gradients into $\reg$ vanish, and FDM rollouts
become insensitive to $\latact$ (register bypass). The pairs remove this
shortcut, since the two flow-matching targets differ only through $\latact$ and
$\latact'$.

The collapse appears early. \Cref{fig:supp-collapse} compares two runs after
2k steps that differ only in whether counterfactual pairs are used, together
with the final model at 13k steps. In the source clip the left arm grasps a
tissue. Without pairs, the FDM sends the arm to the cup whether the latent comes
from this clip or from its still counterfactual: the latent is ignored, and the
model hallucinates a reach for the cup from the first frame alone. With
counterfactual pairs, the still latent already yields a still rollout at 2k
steps, while the moving latent is not yet followed, and at 13k steps both
rollouts match their sources.

%% file: figures/supp_collapse.tex
\begin{figure*}[t]
  \centering
  \newcommand{\cfhead}[1]{{\footnotesize\sffamily #1}}
  \newcommand{\cfok}[1]{{\scriptsize\sffamily\color{teal!70!black}#1}}
  \newcommand{\cfbad}[1]{{\scriptsize\sffamily\color{red!70!black}#1}}
  \newcommand{\cfsrc}[1]{{\scriptsize\sffamily\color{black!75}#1}}
  \setlength{\tabcolsep}{1pt}
  \renewcommand{\arraystretch}{0.4}
  \begin{tabular}{@{}r@{\hspace{4pt}}cc@{\hspace{8pt}}ccc@{}}
    & \multicolumn{2}{c}{\footnotesize\sffamily\bfseries source clip}
    & \multicolumn{3}{c}{\footnotesize\sffamily\bfseries generated end frame} \\[1pt]
    \cmidrule(r{6pt}){2-3}\cmidrule(l{2pt}){4-6}
    & \cfhead{start} & \cfhead{end}
    & \cfhead{\color{red!65!black}\makecell{w/o counterfactual\\pairs, 2k}}
    & \cfhead{\color{teal!70!black}\makecell{with counterfactual\\pairs, 2k}}
    & \cfhead{\color{teal!70!black}\makecell{with counterfactual\\pairs, 13k (final)}} \\[3pt]
    \footnotesize\sffamily\makecell[r]{moving\\source} &
      \raisebox{-0.5\height}{\includegraphics[width=2.95cm]{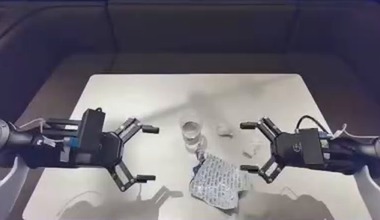}} & \raisebox{-0.5\height}{\includegraphics[width=2.95cm]{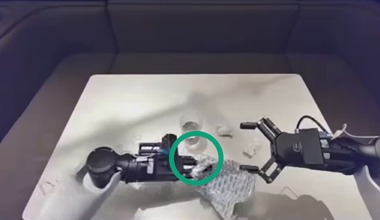}} &
      \raisebox{-0.5\height}{\includegraphics[width=2.95cm]{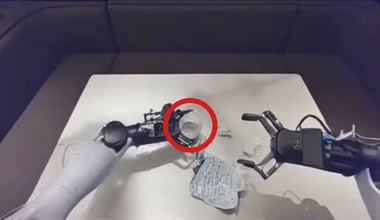}} & \raisebox{-0.5\height}{\includegraphics[width=2.95cm]{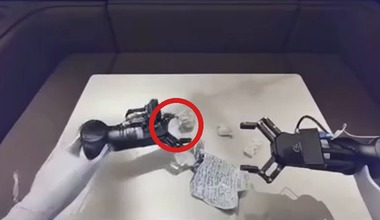}} & \raisebox{-0.5\height}{\includegraphics[width=2.95cm]{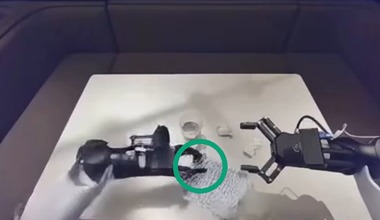}} \\[2pt]
    & & \cfsrc{grasps tissue} & \cfbad{reaches cup} & \cfbad{reaches cup} & \cfok{grasps tissue} \\[4pt]
    \footnotesize\sffamily\makecell[r]{still\\source} &
      \raisebox{-0.5\height}{\includegraphics[width=2.95cm]{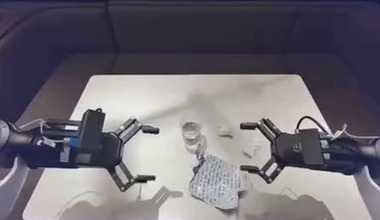}} & \raisebox{-0.5\height}{\includegraphics[width=2.95cm]{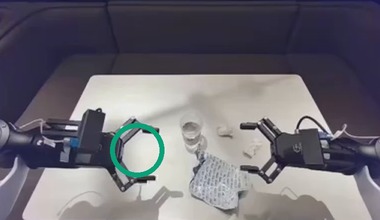}} &
      \raisebox{-0.5\height}{\includegraphics[width=2.95cm]{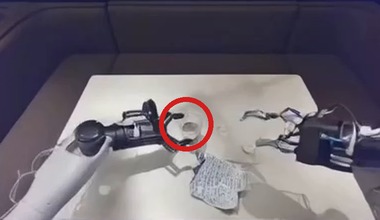}} & \raisebox{-0.5\height}{\includegraphics[width=2.95cm]{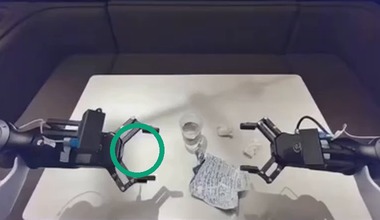}} & \raisebox{-0.5\height}{\includegraphics[width=2.95cm]{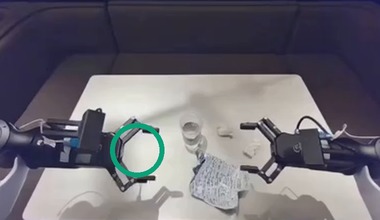}} \\[2pt]
    & & \cfsrc{still} & \cfbad{reaches cup} & \cfok{stays still} & \cfok{stays still} \\
  \end{tabular}
  \caption{\textbf{Counterfactual pairs prevent posterior collapse.} A latent
  action is read in IDM mode from a source clip in which the left arm grasps the
  tissue (top) or from its still counterfactual, which holds the first frame
  (bottom), and the FDM regenerates the clip from the same start frame with
  shared prompt, noise, and seed. Circles mark the end pose of the arm. Trained
  for 2k steps without counterfactual pairs, the FDM reaches for the cup in both
  cases. This is posterior collapse: the FDM ignores the latent and
  hallucinates a reach for the cup from the first frame alone.
  With counterfactual pairs, the same 2k steps already keep the still latent still, although
  the moving latent still reaches for the cup. At 13k steps the generated end
  frames match the source in both cases.}
  \label{fig:supp-collapse}
\end{figure*}

%% file: arxiv/tables/robotwin_realrobot.tex
\begin{table}[t]
  \begin{minipage}[t]{0.40\linewidth}
  \centering
  \caption{\textbf{RoboTwin~2.0 with clean training} (success rate, \%). Policies
  are trained on 50 clean demos per task and tested in clean (c2c) or randomized
  (c2r) scenes, with 100 seeds per task. $\pi_{0.5}$ + past +
  aug.\ uses the same past frame and augmentation as \policyname.}
  \label{tab:robotwin}
  \setlength{\tabcolsep}{5pt}
  \small
  \begin{tabular}{@{}lccc@{}}
    \toprule
    Method & c2c & c2r & Avg. \\
    \midrule
    $\pi_{0.5}$~\cite{black2025pi05}          & 70.66 & 46.00 & 58.33 \\
    $\pi_{0.5}$ + past + aug.              & 81.42 & 55.66 & 68.54 \\
    \policyname\ (ours)                       & \textbf{85.60} & \textbf{62.44} & \textbf{74.02} \\
    \bottomrule
  \end{tabular}
  \end{minipage}\hfill
  \begin{minipage}[t]{0.56\linewidth}
  \centering
  \caption{\textbf{Real-robot tasks} on AgileX NERO. Top: start and end
  of each task. Bottom: success rate (\%) over 30 trials, with 100 training
  demonstrations per task.}
  \label{tab:realrobot}
  \resizebox{\linewidth}{!}{%
  \includegraphics{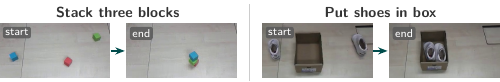}}\\[3pt]
  \setlength{\tabcolsep}{5pt}
  \small
  \begin{tabular}{@{}lcc@{}}
    \toprule
    Method & Stack three blocks & Put shoes in box \\
    \midrule
    $\pi_{0.5}$~\cite{black2025pi05}   & 23.3 & 33.3 \\
    $\pi_{0.5}$ + past + aug.       & 66.7 & 80.0 \\
    \policyname\ (ours)                & \textbf{76.7} & \textbf{90.0} \\
    \bottomrule
  \end{tabular}
  \end{minipage}
\end{table}

%% file: arxiv/tables/robotwin_sota_float.tex
\begin{table}[t]
  \centering
  \caption{\textbf{RoboTwin~2.0 and RoboDojo} (success rate, \%). On RoboTwin,
  methods are trained on 50 clean and 500 randomized demos per task and tested
  in clean (r2c) or randomized (r2r) scenes; other results are from their
  papers or \cite{li2026rotvla,shen2026ld4wam}; GO-1 on RoboTwin is as reported
  in RotVLA~\cite{li2026rotvla}. RoboDojo reports Generalization,
  Precision, Long-Horizon, and Open from the public leaderboard~\cite{chen2026robodojo} as of Nov.~16, 2026 (--: no entry; ours: evaluated by us, mean over three seeds). LA: latent-action pretraining. Best
  \firstbox{bold}, second \secondbox{underlined}.}
  \label{tab:robotwin-sota}
  \small
  \begin{tabular*}{\linewidth}{@{\extracolsep{\fill}}lcc ccc@{\hspace{14pt}\extracolsep{0pt}}c@{\hspace{10pt}}c@{\hspace{10pt}}c@{\hspace{10pt}}c@{}}
    \toprule
    & & & \multicolumn{3}{c}{RoboTwin~2.0} & \multicolumn{4}{c}{RoboDojo} \\
    \cmidrule(lr){4-6}\cmidrule(l){7-10}
    Method & Size & LA & r2c & r2r & Avg. & Generalization & Precision & Long-Horizon & Open \\
    \midrule
    \multicolumn{10}{@{}l}{\textit{World action models}} \\
    Motus~\cite{bi2025motus} & 8B & \checkmark & 88.7 & 87.0 & 87.8 & -- & -- & -- & -- \\
    Being-H0.7~\cite{luo2026beingh07} & 3B & \checkmark & 90.2 & 89.6 & 89.9 & -- & -- & -- & -- \\
    LaWAM~\cite{chen2026lawam} & 2.3B & \checkmark & 92.6 & 89.8 & 91.2 & -- & -- & -- & -- \\
    Fast-WAM~\cite{yuan2026fastwam} & 6B &  & 91.9 & 91.8 & 91.8 & 1.1 & 0.0 & 5.2 & 0.4 \\
    LingBot-VA~\cite{li2026lingbotva} & 5B &  & 92.9 & 91.5 & 92.2 & -- & -- & -- & -- \\
    AHA-WAM~\cite{cai2026ahawam} & 7.2B &  & 93.4 & 92.2 & 92.8 & 3.3 & 2.4 & 2.7 & 0.8 \\
    LD4WAM~\cite{shen2026ld4wam} & 5B & \checkmark & \first{94.0} & 92.8 & 93.4 & -- & -- & -- & -- \\
    \midrule
    \multicolumn{10}{@{}l}{\textit{Vision-language-action models}} \\
    GO-1~\cite{bu2025agibot} & 2B & \checkmark & 37.8 & 36.2 & 37.0 & 1.2 & 0.4 & 0.3 & 0.1 \\
    X-VLA~\cite{zheng2026xvla} & 0.9B &  & 72.8 & 72.8 & 72.8 & 6.8 & 12.0 & 9.8 & 0.5 \\
    $\pi_{0.5}$~\cite{black2025pi05} & 3.3B &  & 82.7 & 76.8 & 79.8 & 8.2 & 5.5 & 14.7 & 1.7 \\
    StarVLA~\cite{starvla2026} & 4B &  & 88.2 & 88.3 & 88.3 & 2.3 & 4.3 & 6.5 & 0.6 \\
    LingBot-VLA~\cite{wu2026lingbotvla} & 4B &  & 88.6 & 86.7 & 87.7 & 4.3 & 1.8 & 5.3 & 0.7 \\
    RotVLA~\cite{li2026rotvla} & 1.7B & \checkmark & 89.6 & 88.5 & 89.1 & -- & -- & -- & -- \\
    HyVLA-0.5~\cite{zhang2026hyembodied} & 4.4B &  & 90.9 & 90.1 & 90.5 & 8.4 & 8.0 & 14.9 & 0.6 \\
    ACE-Ego-0~\cite{li2026aceego} & 4.6B &  & 91.1 & 90.6 & 90.9 & -- & -- & -- & -- \\
    LingBot-VLA~2.0~\cite{wu2026lingbotvla2} & 6B &  & 93.5 & 92.8 & 93.2 & -- & -- & -- & -- \\
    DM0.5~\cite{dexmal2026dm05} & 6B &  & 93.6 & \second{93.3} & \second{93.5} & \second{11.0} & \second{16.8} & \second{19.5} & \second{2.1} \\
    \policyname\ (ours) & 3.8B & \checkmark & \second{93.8} & \first{93.4} & \first{93.6} & \first{17.3} & \first{29.1} & \first{27.0} & \first{2.7} \\
    \bottomrule
  \end{tabular*}
\end{table}

%% file: arxiv/sections/memory.tex
\space \policyname\ observes the current image and a single frame from the segment that
has just ended, whereas the Memory tasks require information that is no longer
visible, and the entries that lead them are built for this purpose; for
example, the DM0.5 team reports that initializing from their memory-pretrained
model improves memory tasks~\cite{dexmal2026dm05memory}. Memory therefore measures a design orthogonal to
our contribution, whereas the other four dimensions measure what latent-action
pretraining is meant to improve: motion that transfers to new scenes, is
accurate, and stays coherent over long tasks.%

%% file: arxiv/tables/efficiency.tex
\begin{table}[t]
  \centering
  \caption{\textbf{Training and inference cost.} Top: dynamics-module training
  cost per step on 33-frame clips ($384{\times}256$, width${\times}$height; batch size 1),
  measured for each module alone and excluding shared attention; values are
  rounded, and those of Kairos~3.1 include a recurrence-equivalent estimate for
  its fused GatedDeltaNet kernels. Bottom:
  median inference latency per control step over 20 runs, measured by us on one
  RTX 6000 Ada, our deployment GPU (BF16, batch size 1, 10 action sampling
  steps, action horizon 32). Each method uses its default input setting:
  the world action models take three views, and \policyname\ takes three views
  and one past frame; text
  and visual conditions are recomputed for every call.}
  \label{tab:efficiency}
  \small\setlength{\tabcolsep}{3pt}
  \begin{minipage}[t]{0.52\linewidth}\centering
  {\sffamily\bfseries Training (per step)}\\[3pt]
  \begin{tabular}{@{}lccc@{}}
    \toprule
    & \makecell{Video expert\\Wan~2.2\\(5B)} & \makecell{Video expert\\Kairos~3.1\\(4.05B)} & \makecell{Latent action\\DiT, ours\\(0.43B)} \\
    \midrule
    Tokens per clip     & 864     & 3{,}456 & 32 \\
    Token dimension     & 3{,}072 & 2{,}560 & 1{,}024 \\
    Forward TFLOPs      & $\approx$7.812 & $\approx$25.651 & $\approx$0.02016 \\
    Backward TFLOPs     & $\approx$15.621 & $\approx$51.294 & $\approx$0.04033 \\
    \bottomrule
  \end{tabular}
  \end{minipage}\hfill
  \begin{minipage}[t]{0.45\linewidth}\centering
  {\sffamily\bfseries Inference (RTX 6000 Ada)}\\[3pt]
  \begin{tabular}{@{}lc@{}}
    \toprule
    Method & Latency (ms) \\
    \midrule
    Motus~\cite{bi2025motus}                       & 900 \\
    Fast-WAM~\cite{yuan2026fastwam} (no VLM) & \\
    \quad joint video prediction & 538 \\
    \quad IDM & 552 \\
    \quad without video generation & 103 \\
    Kairos-WAM~\cite{kairos2026}, w/o video gen. & 194 \\
    \policyname\ (ours)                             & 106 \\
    \bottomrule
  \end{tabular}
  \end{minipage}
\end{table}

%% file: sections/07_conclusion.tex
\section{Conclusion and Future Work}
\label{sec:conclusion}

We show that a pretrained video world model can serve as a native latent action
model. Through dynamics registers, \methodname\ distills motion information
entangled with appearance into a compact latent that transfers across scenes and
embodiments and interpolates into coherent intermediate motions. As a frozen
teacher, it enables \policyname\ to learn from robot, human, and simulated video
without generating video during policy training or inference, improving a
generalist VLA in simulation and on a real robot.

This study establishes the paradigm on Kairos~3.1. Only two of 46 training
sources provide kinematic grounding; the remainder are video-only, and the
evaluation datasets contain no overlapping trajectories, indicating that the
learned action structure is not simply inherited from dense action labels. The
transfer experiments preserve initial geometry while varying embodiment, scene,
or domain, providing a controlled qualitative test of motion retention. Future
work will test other video DiTs, isolate the contribution of kinematic
grounding, and evaluate broader geometric changes and additional robots. As
video world models improve, their internal dynamics may provide an increasingly
scalable foundation for learning actions from video.

%% file: references/bibliography.tex
\begingroup
\small
\bibliographystyle{ieeenat_fullname}
\bibliography{references/references}
\endgroup